\documentclass{article}

\usepackage{arxiv}

\usepackage[utf8]{inputenc}
\usepackage[T1]{fontenc}   
\usepackage{hyperref}      
\usepackage{url}            
\usepackage{booktabs}       
\usepackage{amsfonts}      
\usepackage{nicefrac}       
\usepackage{microtype}      
\usepackage{graphicx}
\usepackage{natbib}
\usepackage{doi}

\usepackage{amsmath, amsfonts, bm}
\usepackage[capitalise,nameinlink]{cleveref}
\crefname{subsection}{Section}{Sections}
\usepackage{tablefootnote, booktabs}
\usepackage{multirow}
\DeclareMathOperator*{\argmin}{argmin}
\usepackage{algorithm}
\usepackage{algorithmic}

\usepackage[usenames,dvipsnames]{color}

\title{Constraint Guided Model Quantization of Neural Networks\thanks{Preprint under review.}}

\author{Quinten Van Baelen\href{https://orcid.org/0000-0003-2863-4227}{\includegraphics[scale=0.06]{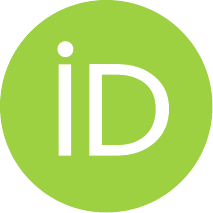}} \\ 
        \texttt{quinten.vanbaelen@kuleuven.be} \\
        KU Leuven, Geel Campus, Dept. of Computer Science; Leuven.AI, B-2440 Geel, Belgium\\
        Flanders Make@KU Leuven, Belgium
       \And
       Peter Karsmakers\href{https://orcid.org/0000-0001-8119-6823}{\includegraphics[scale=0.06]{orcid.pdf}} \\ 
       \texttt{peter.karsmakers@kuleuven.be} \\
       KU Leuven, Geel Campus, Dept. of Computer Science; Leuven.AI, B-2440 Geel, Belgium\\
       Flanders Make@KU Leuven, Belgium}
\date{}

\renewcommand{\headeright}{}
\renewcommand{\undertitle}{}
\renewcommand{\shorttitle}{Constraint Guided Model Quantization of Neural Networks}

\hypersetup{
pdftitle={Constraint Guided Model Quantization of Neural Networks},
pdfsubject={},
pdfauthor={Quinten Van Baelen, Peter Karsmakers},
pdfkeywords={Quantization, Quantization aware training, Memory constrained neural networks, Edge artificial intelligence, Model compression},
}

\begin{document}
\maketitle

\begin{abstract}
	Deploying neural networks on the edge has become increasingly important as deep learning is being applied in an increasing amount of applications. At the edge computing hardware typically has limited resources disallowing to run neural networks with high complexity. To reduce the complexity of neural networks a wide range of quantization methods have been proposed in recent years. This work proposes Constraint Guided Model Quantization (CGMQ), which is a quantization aware training algorithm that uses an upper bound on the computational resources and reduces the bit-widths of the parameters of the neural network. CGMQ does not require the tuning of a hyperparameter to result in a mixed precision neural network that satisfies the predefined computational cost constraint, while prior work does. It is shown on MNIST and CIFAR10 that the performance of CGMQ is competitive with state-of-the-art quantization aware training algorithms, while guaranteeing the satisfaction of an upper bound on the computational complexity defined by the computational resources of the on edge hardware. 
\end{abstract}


\section{Introduction}
Deep neural networks are deployed successfully in many applications \citep{Singh2023}. However, they become increasingly deeper and wider making them require substantial computational resources and memory. Methods that enable reducing the latter while preserving accuracy have perspective to save energy consumption, reduce latency and enable edge AI deployment where only computing hardware with limited resources is available. In the literature many compression techniques, including pruning, quantization, and knowledge distillation, have been proposed for this purpose \citep{Menghani2023}. This work focuses on quantization and proposes the Constraint Guided Model Quantization (CGMQ) method that automatically finds appropriate bit-widths for the model weights and activations such that the memory and computation requirements remain below predefined maxima. Hence a mixed-precision model is targeted where the quantization settings can vary throughout the model. The methods proposed in the literature typically require a tedious iterative process to find appropriate quantization settings. According to \citet{Hohman2024}, targeting a specific model size and complexity cost constraint in one-shot is an important research direction for future work as this reduces the complexity for practitioners to deploy models on the edge. Therefore, the main goal of this work is designing a method that results in a neural network satisfying a given cost constraint without any hyperparameter tuning that is competitive with prior methods with comparable compression rates in terms of performance metrics.

For the task of quantization of a given neural network architecture, there are four classes of methods: (i) gradient based methods \citep{Zhe2019,Lacey2018,Uhlich2020,VanBaalen2020}, (ii) reinforcement learning based methods \citep{Elthakeb2020,Wang2019,Lou2019,Ning2021}, (iii) heuristic based methods \citep{Chu2019,Yao2021,Cai2020,Chen2021}, and (iv) meta-heuristic based methods \citep{Gong2019,Yuan2020,Bulat2021,Wu2018}. For a comprehensive overview the reader is referred to  \cite{Rakka2024}.  \cite{Hohman2024} argue that compressing a network to low bit-width, such as 4 bit, requires compression-aware training. As neural networks are typically trained using gradient descent optimizers, designing a compression method that can be learned with gradient descent allows for an easy combination of both objectives. Hence, the proposed method is a gradient based method, and, therefore, only the gradient based methods are discussed in detail within this work. In \cite{Lacey2018}, the authors proposed a first approach to learn a mixed-precision neural network. A sampling method is proposed to determine during training the bit-width of each layer such that the overall complexity of the neural network is always within some predefined complexity budget. Thus, the training consists of learning the weights of the neural network where the bit-widths are changed accordingly to samples taken from a Gumbel-Softmax distribution and at the end of training the bit-widths are kept fixed for a finetuning step to improve the performance. Therefore, especially in the early stages of training, non-integer bit-sizes are used as a large temperature is used. The temperature parameter is decreased over time such that the samples become closer to integers over time. As a result, a difference between training and inference is observed as the temperature is not set to 0 during training. Furthermore, sampling based methods are expensive \citep{Rakka2024}. 

Differentiable Quantization (DQ) \citep{Uhlich2020} aims at learning two out of the following three: the quantization range, the step size, and the bit-width. The crucial tool that allows the learning of these parameters is the Straight-Through-Estimator (STE) \citep{Bengio2013}. STE ignores the rounding operation during the backward pass, resulting in the removal of the zero gradients of the step-function defined by the rounding operation by definition. Moreover, a custom gradient has been defined to learn two of the quantization range, the step size, and the bit-width, since the third can be inferred from the other two. Within DQ, three constraints are considered: i) the total memory to store all weights should be below some threshold, ii) the total memory to store all feature maps should be below some threshold, and iii) the memory required to store the largest feature map should be below some threshold. These three constraints are added to the learning objective as a regularization term as is done in the penalty method \citep{Bertsekas2014}. As mentioned by \cite{Uhlich2020} this approach does not guarantee the constraints to be satisfied. Manual tweaking of the hyperparameters of the regularization is required to increase the likelihood of the constraints to be satisfied.

The Bayesian Bits (BB) \citep{VanBaalen2020} introduces the idea of gate variables. BB considers quantization that is a power-of-two, which is required in many hardware devices \citep{Nagel2021}. The learning of BB uses variational inference by defining a prior that gives a larger penalty to higher bit-widths. However, the resulting regularization term is an expectation of a log-likelihood of the stochastic gate variables. This is expensive to compute and, therefore, the reparametrization trick \citep{Kingma2013} is used. The stochastic gates during training are replaced by deterministic gates during inference. Note that this results in possibly different predictions obtained for a single sample when the network is in training mode and when the network is in inference mode. Similarly as for DQ, no guarantee can be obtained for a possible computational cost constraint. Again, a hyperparameter, related to the regularization term, can be iteratively modified to meet finally the predefined cost constraint. 

BATUDE \citep{Yin2022} aims at determining the Tucker rank of the weights such that the cost constraint is satisfied. However, solving this problem exactly is NP-hard \citep{Hillar2013}. Therefore, BATUDE considers a relaxation of the original problem, which loses the possible guaranteed satisfaction of the cost constraint. This relaxation is obtained by considering the Tucker-2 nuclear norm as a regularization term. The resulting constrained optimization problem only constraints the cost constraint and is solved by a budget constrained augmented direction Lagrangian, which is a special case of the Douglas-Rachford splitting algorithm \citep{Gabay1976,Eckstein1992}.

\cite{Verhoef2019} propose to train quantized neural networks iteratively by lowering the bit-widths of the weights gradually. This method leads to a single bit-width for all weights and multiple training cycles. In particular, multiple choices exist to arrive to a low bit-width network, for example, in order to arrive at a quantized network with 4 bit weights, a quantized network with 16 bit weights can be considered alone or also a quantized network 8 bit weights. This leads to the tuning of the considered bit widths. A disadvantage of this approach is that larger steps in complexity can lead to a larger drop in performance, which cannot easily be resolved. Furthermore, in this work, it is not possible to assign higher bit-widths to some parts of the network such as part of the feature extraction.

myQASR \citep{Fish2023} uses a small set of unlabelled examples to find the bit-widths of each layer of a neural network. The method is build on their observation that a positive correlation exists between the median of activations and the quantization error. Therefore, myQASR computes the median and reduces the bit-width of the layer with the smallest absolute value of the median of the activations by 1. This process is repeated by considering the layer with the largest bit-widths and until the cost constraint is satisfied. After the cost constraint is met, the bit-width are kept fixed and the network is trained for improving the performance. This results in a network with at most 2 different bit-widths. 

The proposed method aims at learning the bit-widths of a mixed precision quantized neural network in combination with learning of the quantization ranges for a power-of-2 quantization scheme and the weights and the biases of a neural network. In particular, the main contributions are:
\begin{itemize}
    \item A one-shot method CGMQ that allows to obtain a quantized version of a pre-trained neural network model that satisfies a cost constraint defined by the computational resources of the device on which the model will be deployed. Furthermore, the method is independent of the chosen quantization method and does not alter the internals of the network during training nor inference. 
    \item The proposed method is validated on MNIST and CIFAR10, two standard benchmarks for quantization methods for neural networks. It is shown that CGMQ is competitive with prior work with respect to the performance metrics, while guaranteeing the satisfaction of the cost constraint.
\end{itemize}

The remainder of this text is structured as follows. First, the proposed method Constraint Guided Model Quantization is defined in \cref{sec:CGMQ}. Afterwards, in \cref{sec:TheoreticalComparison} a qualitative comparison with prior work is made. Next, an experimental evaluation is discussed in \cref{sec:Experiments} with an ablation study in \cref{sec:AblationStudy}. Possible directions for future work are described in \cref{sec:FutureWork}. Finally, the text is concluded in \cref{sec:Conclusion}.

\section{Constraint guided model quantization}\label{sec:CGMQ}
The proposed method CGMQ targets solving the following optimization problem
    \begin{align*}
        \argmin \limits_{\theta^{(q)}} \quad & L\left( \bm{\Phi} \left(\bm{x} \mid \theta^{(q)}\right), \bm{y} \right) \\
        \textrm{s.t.} \quad & \textrm{cost}\left(\bm{\Phi} \left(\cdot \mid {\theta^{(q)}}\right)\right) \leq B_{\textrm{BOP}},
    \end{align*}
where $\bm{\Phi}\left(\cdot\mid\theta^{(q)}\right)$ is the quantized model with quantized weights $\theta^{(q)}$ and bit-widths $q$, $\bm{x}$ is a batch of input samples, $\bm{y}$ is a batch of groundtruth labels corresponding to $\bm{x}$, $L$ the loss function, $\textrm{cost}$ a function that computes the cost of a quantized model, and $B_{\textrm{BOP}}$ the maximally allowed cost of the quantized model $\bm{\Phi}\left(\cdot\mid\theta^{(q)}\right)$.

Although different cost functions ($\textrm{cost}(\cdot)$) can be defined for which the proposed optimization approach is suited, in this work it is opted to relate it to the number of Bit-Operations (BOP) which is a hardware agnostic proxy to model complexity. The BOP metric takes into account the number of required FLOPs to calculate a model's output and the bit-widths of weights and activations. Note for simplicity the term \textit{weight} is used to refer to all learnable parameters (hence also bias), except when specified explicitly otherwise. To compute the BOP count, the definition from \cite{Uhlich2020,Baskin2018} is adopted. More details are added in \cref{ssec:BOPcost}. Beforehand, an upper bound on the BOP cost $B_{\textrm{BOP}}$ is defined by the practitioner. Next a quantized model with proper weight and activation bit-widths is searched for that optimizes the minimization objective while satisfying the cost (BOP) constraint.

The remainder of this section is structured as follows. In \cref{ssec:GatingWeightsAndActivations}, the gate variables that determine the bit-width of the weights are defined. Afterwards, the core component of CGMQ, which is the algorithm used for learning the gate variables, is introduced formally in \cref{ssec:LearningGateVariables}. These two sections are generic and can be adopted by future work. The different directions considered in this work are defined in \cref{ssec:DirectionOfGateVariables}. The input of CGMQ is not a neural network with random weights. Hence, \cref{ssec:ModelQuantizationIntialization} discusses how the initialization of CGMQ can be obtained. The computations used to compute the BOP count of a given quantized neural network are discussed in \cref{ssec:BOPcost}. At last, the full algorithm is is summarized in \cref{ssec:Algorithm}.

\subsection{Gating weights and activations}\label{ssec:GatingWeightsAndActivations}
In order to automatically determine appropriate bit-widths for all weights, and activations an auxiliary gate variable $g$ is introduced. The gate variable of a weight $w$ is denoted by $g_w$ and the gate variable of a neuron is denoted by $g_a$ as it corresponds to the value of the neuron after the activation. When $g$ is used it refers to both $g_w$ and $g_a$. The value of the gate variable $g$ is used to determine the bit size of the corresponding weight or activation. For this purpose, the weight and activation values are decomposed similar to the approach in \citep{VanBaalen2020}. Hence, a power-of-2 quantization scheme is considered as this can efficiently be implemented in hardware \citep{Nagel2021}, even though this is not required for CGMQ. Before revisiting this decomposition, first the quantization function $x_b = Q(x,b,\alpha, \beta)$, that quantizes a floating point value in the range $\{x | x \in \mathbb{R}, x \in [\alpha, \beta]\}$ to a quantized value that only uses $b$ bits, is defined by
    \begin{equation}\label{eq:FakeQuantization}
        Q(x,b,\alpha,\beta) := \frac{\beta-\alpha}{2^b -1} \left\lfloor \frac{\textrm{clip}_{[\alpha,\beta]}\left(x\right)\left(2^b -1\right)}{\beta-\alpha}\right\rceil,
    \end{equation}
where $\left\lfloor \cdot \right\rceil$ denotes the round-to-nearest-integer function, and $\textrm{clip}_{[\alpha, \beta]}(x)$ is the clipping function of $x$ to the interval $[\alpha, \beta]$, which is defined by
    \begin{equation*}
        \textrm{clip}_{[\alpha,\beta]}:\mathbb{R}\to\mathbb{R}: x\mapsto \begin{cases}
            \alpha, & \textrm{if } x < \alpha, \\
            x, & \textrm{if } x\in[\alpha, \beta], \\
            \beta, & \textrm{if } x>\beta.
        \end{cases}
    \end{equation*}
    
\begin{figure}[ht]
    \centering 
    \includegraphics[scale=0.4]{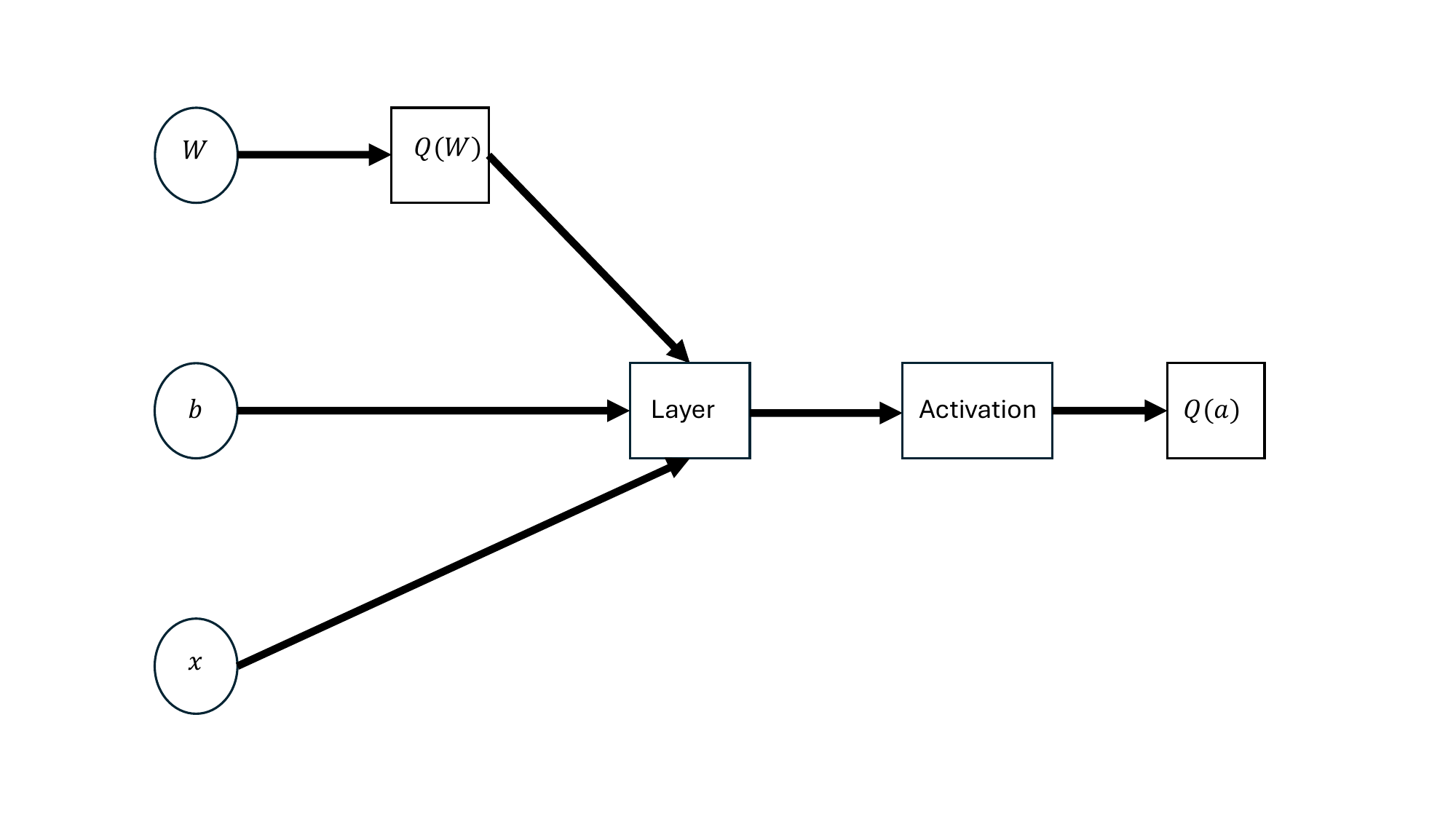}
    \caption{Illustration of fake quantization of a layer, where the input ($x$) is assumed to be quantized.}
    \label{Fig:QuantizationLayer}
\end{figure}

The quantization is performed on the weights and the output or activation of a given layer. This procedure is called Fake Quantization (FQ) as it emulates the quantization behavior while all variables are still in floating point. In this way a highly dynamic range is obtained which is convenient for learning. A schematic illustration is given in \cref{Fig:QuantizationLayer}, where $W$ is the weight tensor, $b$ the bias tensor, $x$ the input of the layer, $Q(W)$ the quantization of a weight tensor $W$ with the ranges omitted for ease of notation, Layer denotes the type of layer, Activation denotes the activation function of the layer, and $Q(a)$ is the quantization of the output or activation of the layer with the ranges omitted for ease of notation. Furthermore, $\alpha$ is defined as $-\beta$ if the tensor that needs to be quantized contains (strictly) negative values and $\alpha$ is defined as $0$ if all values in the tensor that needs to be quantized are positive.

In general the notation $x_b$ indicates that it is the $b$-bit quantization of $x$. As is explained by \cite{VanBaalen2020}, this quantization process can be decomposed as given by the following equation
    \begin{equation}\label{def:RecursiveQuant}
        x_{32} := x_2 + \sum\limits_{j \in \mathcal{B}} \varepsilon_{j},
    \end{equation}
where in our case $\mathcal{B} = \left\{4,8,16,32\right\}$ as these power of $2$ bit-widths in computing hardware lead to more efficient computations \citep{Nagel2021}, and $\varepsilon_j := x_j - x_{2^{log_2(j)-1}}$ the residual quantization error when $j$ instead of $2^{log_2(j)-1}$ bits are used. Now Equation \ref{def:RecursiveQuant} can be modified by adding an auxiliary binary gate function $G_i$, which is applied to each gate variable $g$, as follows
    \begin{equation}\label{def:quantDecomp}
        x_b = G_2(g) \left[ x_2 + G_4(g) \left[ \varepsilon_4 + G_8(g) \left[ \varepsilon_8 + G_{16}(g) \left[ \varepsilon_{16} + G_{32}(g) \,\, \varepsilon_{32} \right] \right] \right] \right].
    \end{equation}
This auxiliary gate function is defined as
    \begin{equation*}
            G_b:\mathbb{R} \to \{0,1\}: g \mapsto \begin{cases}
                1\text{,} & \text{if } T(g) \geq b\text{,} \\
                0\text{,} & \text{otherwise,}
            \end{cases}
    \end{equation*}
with transformation function $T$ equal to
    \begin{equation} \label{eq:DefT}
        T:\mathbb{R} \to \mathbb{R}: g \mapsto \begin{cases}
            0\text{,} & \text{if }g\leq 0\text{,} \\
            2\text{,} & \text{if }g\in\left(0, 1\right]\text{,} \\
            4\text{,} & \text{if }g\in\left(1, 2\right]\text{,} \\
            8\text{,} & \text{if }g\in\left(2, 3\right]\text{,} \\
            16\text{,} & \text{if }g\in\left(3, 4\right]\text{,} \\
            32\text{,} & \text{if }g > 4\text{.}
        \end{cases}
    \end{equation}
Observe that when $g$ has a value resulting in a bit-width of $4$, e.g. when $g=1.5$, it holds that $G_2(g)=1$, $G_4(g)=1$, $G_8(g)=0$, $G_{16}(g)=0$, and $G_{32}(g)=0$. Therefore, for this example it follows that $x_4 = x_2 + \varepsilon_4$ or equivalently that $x_4$ has indeed a bit-width of $4$. In this work, pruning is not yet considered and left as future work. Therefore, as soon as a value $g<0.5$ is obtained, it is replaced with $0.5$.

The gate variable can correspond to a single neuron or weight, or to a set of neurons and/or weights. For example, one gate variable can control the bit-width of all weights in a single layer. This is a choice that can be made depending on the desired application. Moreover, it allows to use a given hardware device in an optimal manner. In this work, every single weight and every single activation has a gate variable. We follow the common approach \citep{Krishnamoorthi2018} of not quantizing biases and learning the bit-widths of weights and activations as biases require high bit-widths in good performing models.

\subsection{Learning of gate variables}\label{ssec:LearningGateVariables}
Before describing in detail how the gate variables are learned by CGMQ, Quantization Aware Training (QAT) \citep{Jacob2017,Verhoef2019,Nagel2021} is briefly reviewed as this forms the basis of CGMQ. In QAT, FQ blocks are inserted into different parts of the neural network architecture when it is trained. The FQ is done for the weights and the activations. This makes the network already aware of precision limitations during training and hence makes it more robust to these effects. The FQ blocks have an input-output relation as is defined by Equation \ref{eq:FakeQuantization}. They are used to quantize float values into fixed point alternatives given some predefined bit-width. 

During training gradients need to be calculated. However, the round-to-nearest-integer function $\left\lfloor \cdot \right\rceil$ within a FQ block, is a non-linear function, whose gradient is almost everywhere equal to 0. This prohibits proper learning. In order to enable learning the straight-through estimator \citep{Bengio2013} is used. This allows using the round-to-nearest-integer during the forward-pass while the identity function is used in the backward pass, leading to a non-zero gradient for the round-to-nearest-integer operator during training.

In standard QAT the bit-width is set beforehand. Therefore, the behavior of the FQ blocks remains the same throughout the full learning process. In CGMQ the bit-width of each FQ block is controlled by its corresponding gate variable $g$ as its value will quantize a float using Equation \ref{def:quantDecomp}.

Given an already trained model with floating point weights and activations, CGMQ selects the most appropriate bit-widths by changing the values for gate $g$. In Equation \ref{def:quantDecomp} the non-linear functions $G_b$, as the round-to-nearest-integer function does in the FQ blocks, will make the gate variable $g$ to not have a non-zero gradient. However, CGMQ resolves this issue by defining a ``direction'' $dir$ which is used as a gradient, although it is not a gradient, in a gradient descent optimizer to have it use the following update step
    \begin{equation*}
        g^{(k+1)} = g^{(k)} - \eta_g \,\, dir, 
    \end{equation*} 
where $k$ indicates the iteration number, and $\eta_g$ the learning rate.

\subsection{Direction of gate variables}\label{ssec:DirectionOfGateVariables}

The direction $dir$ should satisfy the following two properties: (i) if the cost constraint is not satisfied, then the direction should be (strictly) positive, (ii) if the cost constraint is satisfied, then the direction should be negative or $0$. These two properties are a consequence of the observation that increasing the bit-width of variables cannot have a negative influence on the loss of the network, after possibly adjusting the weights of the network, as the network can represent more functions as long as the network does not overfit. However, the loss value on the training set can only improve when the bit-width are increased. Hence, the bit-width of a variable should only be decreased when the cost constraint is violated as this reduces the complexity of the neural network both in terms of the class of functions that it can represent as well as its cost (in terms of BOP). Observe that a lower complexity can improve the latency or improve the robustness of the network, but the \textit{optimal} set of bit-widths such that the network satisfies the cost constraint should be learned from data. The first property for the direction will make sure that each gate variable present in the network will become smaller after applying a gradient descent update. Therefore, if this direction is sufficiently large in absolute value and the learning rate is kept constant, the bit-width will decrease. In other words, the cost constraint will be satisfied if sufficient optimization iterations are performed. The second property allows to grow the gate variable of different weights at different rates. Hence, not all gate variables can pass a threshold such that they result in a different bit-width. In particular, it follows that the second property allows for adjusting the bit-width of a single variable if not all gate variables are updated with the same direction. Indeed, a gate variable with a smaller direction, as the direction is negative in this case, is expected to lead to an increase in bit-width faster. As a consequence, the training allows to deviate from the first allocation of bit-widths that satisfy the cost constraint. In particular, if the direction is a function that depends on the loss function, this could lead to a better performing model. The combination of these two properties allows the change of two $16$-bit variables and one $2$-bit variable into one $16$-bit variable and two $8$-bit variables if the cost constraint is formulated as a maximal number of $36$-bits is allowed. In general, if there exist real numbers $K_1, K_2>0$ and $K_3, K_4<0$ such that the direction is in $[K_1, K_2]$ when the cost constraint is not satisfied and in $[K_3, K_4]$ when the cost constraint is satisfied, then a desired behavior is obtained for gradient descent for each gate variable. 

In this work, three methods to determine the appropriate $dir$ values are proposed. However, it should be stressed that any method that results in a number that can be used by a gradient descent optimizer can be used as long as the two properties above are satisfied.

The first method ($dir_1$) uses the idea that decreasing the bit-width of weights that have a small gradient of the loss function in absolute value is expected to have a smaller impact on the resulting prediction compared to decreasing the bit-width of weights that have a large loss gradient in absolute value. Therefore, gate variables $g_w$ are updated using $dir_1^{(w)}$ which is defined as
    \begin{equation*}
        dir_1^{(w)}: \left\{Sat, Unsat\right\}\to \mathbb{R}: s\mapsto \begin{cases}
            \dfrac{1}{\dfrac{1}{N_b}\left| \sum\limits_{i=1}^{N_b} \nabla_w L\left(\bm{\Phi}\left(x_i\mid\theta^{(q)}\right),y_i \right) \right|}, & \text{if $s=Unsat$,} \\[1.25cm]
            -\left|g_w\right|, & \text{if $s=Sat$,}
        \end{cases}
    \end{equation*}
where $s=Unsat$ if the cost constraint is unsatisfied, that is, in this case when $\textrm{cost}\left(\bm{\Phi}\left(\cdot\mid\theta^{(q)}\right)\right) > B_{\textrm{BOP}}$, $s=Sat$ if the cost constraint is satisfied, $\nabla_w L\left(\bm{\Phi}\left(x_i\mid \theta^{(q)}\right),y_i\right)$ is the partial derivative of $L\left(\bm{\Phi}\left(x_i\mid \theta^{(q)}\right),y_i\right)$ to $w$, and $N_b$ the mini-batch size. When the cost constraint is satisfied, then the bit-width is increased with the absolute value of the gate variable. This corresponds to making it more likely to increase the bit-width of weights with a large bit-width compared to weights with a small bit-width. As the bit-width of all weights are decreased the fastest for weights with a small gradient, then this proposed increase will result in weights with a small gradient having a small bit-width and weights with a large gradient having a large bit-width.

Similarly, for each activation $a$, its corresponding gate variable $g_a$ is adapted via 
    \begin{equation*}
        dir_1^{(a)}:\left\{Sat, Unsat\right\}\to\mathbb{R}:s\mapsto \begin{cases}
            \dfrac{1}{\dfrac{1}{N_b}\left|\sum\limits_{i=1}^{N_b} \nabla_a L \left(\bm{\Phi}\left(x_i\mid\theta^{(q)}\right),y_i\right) \right|}, & \text{if $s=Unsat$,} \\[1.25cm]
            -\left|g_a\right|, & \text{if $s=Sat$,}
        \end{cases}
    \end{equation*}
where $s=Unsat$ if the cost constraint is unsatisfied, that is, in this case when $\textrm{cost}\left(\bm{\Phi}\left(\cdot\mid\theta^{(q)}\right)\right) > B_{\textrm{BOP}}$ and $s=Sat$ if the cost constraint is satisfied.

The second method ($dir_2$) uses the idea that not only the gradient of a weight determines how much the overall prediction is dependent on the weight but the absolute value of the weight as well. Similarly, when the cost constraint is satisfied, the gate variable is increased with the absolute value of the weight and the absolute value of the gate variable such that large weights are expected to have larger bit-widths. Furthermore, the weights with a large gate variable can be assigned a higher bit-width faster to allow weights to have vastly different bit-widths. Moreover, the mean is taken over the batch similarly as before. Therefore, the gate variables $g_w$ are updated using $dir_2^{(w)}$ which is defined as
    \begin{equation*}
        dir_2^{(w)} :\left\{Sat, Unsat\right\} \to\mathbb{R} :s\mapsto \begin{cases}
            \dfrac{1}{\dfrac{1}{N_b} \left|\sum\limits_{i=1}^{N_b} \nabla_w L\left(\bm{\Phi}\left(x_i\mid \theta^{(q)}\right), y_i\right) \right| + \left| w\right|}, & \\ & \phantom{\text{if $s=Sat$.}}\llap{\text{if $s=Unsat$,}} \\
            -\left(\left|g_w\right| + \left|w\right| \right), & \text{if $s=Sat$.}
        \end{cases}
    \end{equation*}
Similarly, for each activation $a$, its gate variable $g_a$ is adapted via
    \begin{equation*}
        dir_2^{(a)}:\left\{Sat, Unsat\right\} \to\mathbb{R}:s\mapsto \begin{cases}
            \dfrac{1}{\dfrac{1}{N_b} \left[\left| \sum\limits_{i=1}^{N_b} \nabla_a L\left( \bm{\Phi}\left(x_i\mid\theta^{(q)}\right), y_i\right) \right| + \left|\sum\limits_{i=1}^{N_b} \bm{\Phi}_a\left(x_i\mid\theta^{(q)}\right) \right| \right]}, \\[1.25cm] & \phantom{}\llap{\text{if $s=Unsat$,}} \\
            -\left(\left|g_a\right| + \dfrac{1}{N_b} \left|\sum\limits_{i=1}^{N_b} \bm{\Phi}_a\left(x_i\mid\theta^{(q)}\right) \right|\right), & \llap{\text{if $s=Sat$,}}
        \end{cases}
    \end{equation*}
where $\bm{\Phi}_a\left(x_i\mid \theta^{(q)}\right)$ is the value of activation $a$ for input $x_i$.

Finally, the third method ($dir_3$) is based on the first order Taylor approximation, meaning that both the absolute value of a weight and the absolute value of the gradient with respect to this weight are used to adjust the gate variable. This uses the observation that a weight with large absolute value can also have a large influence on the predictions even though the partial derivative is small in absolute value. In other words, gate variables $g_w$ are updated using $dir_3^{(w)}$ which is defined as
    \begin{equation*}
        dir_3^{(w)}:\left\{Sat, Unsat\right\}\to\mathbb{R}:s\mapsto \begin{cases}
            \dfrac{1}{\dfrac{1}{N_b} \left|\sum\limits_{i=1}^{N_b} \nabla_w L\left(\bm{\Phi}\left(x_i\mid\theta^{(q)}\right), y_i\right)\right| + \left|w\right|}, & \\[1.25cm] &\phantom{}\llap{\text{if $s=Unsat$,}} \\[0.5cm]
            -\left(\dfrac{1}{N_b} \left|\sum\limits_{i=1}^{N_b} \nabla_w L\left(\bm{\Phi}\left(x_i\mid\theta^{(q)}\right), y_i\right) \right| + \left|w\right|\right), & \\[0.5cm] &\phantom{}\llap{\text{if $s=Sat$.}}
        \end{cases}
    \end{equation*}
    
Similarly, for each activation $a$, its gate variable $g_a$ is adapted via
    \begin{equation*}
        dir_3^{(a)}:\left\{Sat, Unsat\right\} \to\mathbb{R}:s\mapsto \begin{cases}
            \dfrac{1}{\dfrac{1}{N_b}\left[\left| \sum\limits_{i=1}^{N_b} \nabla_a L\left( \bm{\Phi}\left(x_i\mid\theta^{(q)}\right), y_i\right) \right| + \left|\sum\limits_{i=1}^{N_b} \bm{\Phi}\left(x_i\mid\theta^{(q)}\right) \right| \right]}, \\[1.25cm] & \phantom{}\llap{\text{if $s=Unsat$,}} \\[0.5cm] 
            \dfrac{-1}{N_b} \left( \left|\sum\limits_{i=1}^{N_b} \nabla_a L\left(\bm{\Phi}\left(x_i\mid\theta^{(q)}\right), y_i\right) \right| \right. \\ \left.\qquad\qquad\qquad\qquad + \left|\sum\limits_{i=1}^{N_b} \bm{\Phi}\left(x_i\mid\theta^{(q)}\right) \right|\right), & \\[0.5cm] & \phantom{}\llap{\text{if $s=Sat$,}}
        \end{cases}
    \end{equation*}

All choices above for $dir$ do not require to obtain a non-zero gradient for the gate variables themselves. Hence, the step function which is used for transforming the gate variable into the corresponding bit-width does not need to be adjusted during training. 

\subsection{Model quantization initialization}\label{ssec:ModelQuantizationIntialization}

As mentioned earlier, the input of CGMQ is not a neural network with randomly initialized weights. The input of CGMQ is obtained in the following steps, which is similar to the standard approach for quantization aware training \citep{Nagel2021}. First, the neural network is pre-trained in floating point, here the network is trained for 250 epochs. Second, the quantization ranges are calibrated when the model uses FQ and all weights are set to a bit-width of 32. The calibration of the quantization ranges is done differently for the weights and the activations. The quantization range of the weights is obtained by computing the maximum and the minimum of the weights for each layer individually. The batch normalization layers are handled by unfolding them \citep{Krishnamoorthi2018}. Therefore, as weights are not updated during calibration of the quantization ranges, the quantization range of the weights is constant. When all weights of a layer are positive, then $\beta$ is set to this maximum and $\alpha$ is set to 0. When not all weights of a layer are positive, then $\beta$ is set to the maximum and $\alpha=-\beta$. For the activations, a similar approach is used except that a running mean is used to update the ranges. The momentum of this running mean is 0.1. Third, the quantization ranges are learned for 20 epochs in order to improve the performance of the model with 32 bit-width weights and activations. The resulting model is used as input to the CGMQ method.

\subsection{BOP cost}\label{ssec:BOPcost}
The BOP cost is computed as follows. For a given layer, the BOP count is given by the sum over all activations of the product of the bit-width of the activation with the sum of the bit-widths of the weights that determine the activation. In other words, for a dense layer $l$ the BOP count is
    \begin{equation*}
        BOP(l) = \left< \sum\limits_{j} b_{\bm{W}_{i,j}}, b_{\bm{a}} \right>,
    \end{equation*}
where $b_{\bm{W}_{i,j}}$ is the matrix of bit-widths of the weights of the layer $l$, $b_{\bm{a}}$ is the vector of bit-widths of the activations of the layer $l$, $\left<\cdot,\cdot\right>$ is the standard scalar product, and the convention is followed that the dense layer is defined as $l(x):= \bm{W}^T x + \bm{a}$. For a convolutional layer, this results in the sum over all activations of the product of the bit-width of the activation and the sum of all the bit-widths in the filter corresponding activation. Observe that in case all learnable parameters of a single channel have the same bit-width that this reduces to the BOP count used by \cite{VanBaalen2020}. The satisfaction of the cost constraint defined by an upper bound on the total BOP count is only checked at the end of the epoch and this result is used to determine the case of $dir$ during the next epoch.

\subsection{The algorithm}\label{ssec:Algorithm}

The initialization of the CGMQ method is a pre-trained 32 bit-width quantized neural network, as is standard for quantization aware training methods. In this work, this is obtained by training the network in floating point, calibration of the quantization ranges with all variables in 32 bit-width, and fine-tuning the quantization by learning them.

Given this initialization, CGMQ aims at learning weights, biases, quantization ranges, and the gate variables all combined. A pseudo-code is provided in \cref{alg:CGMQPseudoCode}. The training procedure defined by CGMQ updates the gate variables in every single training iteration. Therefore, it is possible that in every single training step the bit-width of some weights are adjusted. This adjustment is hard for the neural network as this difference in bit-width can be understood as an adjustment in architecture. Hence, we propose to alternate between epochs in which CGMQ is applied and epochs where the weights and quantization ranges are learned for fixed gate variables. As a result, the network can more easily adjust its weights and quantization ranges to cope with the difference in bit-widths and yielded better results.

\begin{algorithm}[ht]
	\caption{CGMQ}
	\label{alg:CGMQPseudoCode}
	\textbf{Input:} batches of training examples $\{\bm{x}_i\}_i$ with ground-truth labels $\{\bm{y}_i\}_i$, pre-trained network $\bm{\Phi}(\cdot\mid \theta^{(q)})$, loss function $L$, number of epochs $E$, learning rates for trainable variables $\{\eta_{\theta^{(q)}}, \eta_g\}$, cost function $\textrm{cost}$ and upper bound on cost constraint $B_{\textrm{BOP}}$
	\begin{algorithmic}[1]  
        \FOR {$epoch=1:E$}
        \IF {$\textrm{cost}\left(\bm{\Phi} \left(\cdot\mid \theta^{(q)} \right) \right) \leq B_{\textrm{BOP}}$}
            \STATE {$s=Sat$}
        \ELSE
            \STATE {$s=Unsat$}
        \ENDIF
        \FOR {$\bm{x} \in \{\bm{x}_i\}_i$}
        \STATE {Compute $L\left(\bm{\Phi}\left(\bm{x}_i\mid\theta^{(q)}\right) , \bm{y}_i\right)$}
        \STATE {$\theta^{(q)} \leftarrow \theta^{(q)} - \eta_{\theta^{(q)}} \nabla L\left(\bm{\Phi}\left(\bm{x}_i\mid\theta^{(q)}\right) , \bm{y}_i\right)$\\$g\leftarrow g - \eta_g dir(s)$}
        \ENDFOR
        \ENDFOR
	\end{algorithmic}
\end{algorithm}

\section{Qualitative comparison}\label{sec:TheoreticalComparison}

A first crucial property of CGMQ does not yield different predictions when a fixed neural network is in training mode or inference mode with respect to the quantization. Of course, if the network contains batch normalization layers, then there is a difference in predictions when the model is in training mode or in inference mode, but this is the same for a floating point model. In other words, applying CGMQ yields no additional differences between training and inference mode compared to the floating model. Not all state-of-the-art methods have this property as this often yields a gradient of 0 with respect to the bit sizes. For example, the BB method \citep{VanBaalen2020} does alter the forward pass.

Second, the BB method requires to store more gate variables compared to CGMQ. The proposed method uses a single variable to determine the bit-width of a single weight and BB uses 5 variables to determine the bit-width. In particular, if pruning is considered, CGMQ uses still a single variable, while in this case BB uses 6 variables. Therefore, CGMQ can train a larger model on a given computing platform as the additional computational and memory requirement are smaller compared to BB. Furthermore, CGMQ is not based on reinforcement learning nor on any stochastic process. Hence, CGMQ is generally more efficient as argued by \cite{Uhlich2020}.

Third, CGMQ supports to train all the different variables together, that is, the weights, the quantization ranges, and the bit sizes can be learned all combined. Moreover, the general idea can be applied to other choices of FQ, for example by adding a translation to the quantization function in Equation \ref{eq:FakeQuantization}, as any direction that satisfies the two desired properties listed in \cref{ssec:LearningGateVariables} yield the desired result since the gradient of the loss function with respect to the gate variables is zero. 

Fourth, applying quantization algorithms is difficult for practitioners \citep{Pandey2023}. CGMQ improves on this aspect as it does not require the fine-tuning of a hyperparameter in order to satisfy a cost constraint, which makes it easier for practitioners to apply quantization algorithms compared to other methods that consider a computational cost constraint. Of course, the proposed method does not alleviate all possible difficulties of running a quantization algorithm. 

Finally, CGMQ as presented in this work, guarantees that some model is found that satisfies the cost constraint as long as such a model exists. Indeed, by using the satisfaction of the cost constraint at the end of the epoch to determine the corresponding case of $dir$, it follows that the cost decreases during an epoch if the cost constraint is satisfied. Moreover, as all choices for $dir$ are non-zero, the gate variables will keep on decreasing until the cost constraint is satisfied at the end of the epoch. However, at this point in training a model is found that satisfies the cost constraint. The only method that considers a similar constraint is DQ \citep{Uhlich2020}. However, in DQ no guaranteed satisfaction of the constraint can be obtained as the constraint is added to the unconstrained objective as a regularization term.

\section{Experimental setup and results}\label{sec:Experiments}

Two sets of experiments are performed. The first set of experiments compares CGMQ to state-of-the-art methods where bound in the cost constraint that is used for CGMQ that is close to the obtained BOP count of the other methods. The second set of experiments considers different bounds on the cost constraint for CGMQ and the resulting performance.

\subsection{Data sets and model architectures}
The performance of CGMQ is compared with prior work for different bounds on the cost of the network on MNIST and CIFAR10 as is often done in quantization on small data sets. The network on MNIST is a LeNet-5 and on CIFAR10 is a VGG-7 as is done by \citep{Liu2016}. 

The MNIST data set is preprocessed by normalizing each image to have 0.5 as mean and 0.5 as standard deviation. The training set of CIFAR10 is preprocessed by performing random cropping, randomly flipping horizontally and a normalization along the channels such that the three channels have as mean $(0.4914, 0.4822, 0.4465)$ and as standard deviation $(0.2023, 0.1994, 0.2010)$. The test set of CIFAR10 has the same normalization but the random cropping and flipping is not considered. Similar to the implementation of BB, the batch normalization layers use batch statistics also during inference time. We observed a drop in performance when the running mean of the average and the standard deviation are used on training and test set. Hence, a similar approach to BB is used in the experiments.

\subsection{Configuration of learning methods}

The weights and quantization ranges are learned using Adam \citep{Kingma2015}. For CGMQ, both individual gate variables as a single gate variable for each layer is used. The latter is for a fair comparison with prior work. The gate variables are learned with a standard gradient descent optimization without momentum. The learning rate for the weights and the quantization ranges is set to $0.001$. On MNIST, the learning rate for the gate variables is set to $0.01$ for $dir_1$, $dir_2$, and to $0.001$ for $dir_3$. These choices for the learning rates of the gate variables yielded the best results. A smaller learning rate for $dir_3$ is most likely a consequence of the fact that by taking into account the magnitude of the weights as well as the magnitude of the gradient of the weights, the resulting gradient is significantly larger compared to the other methods. On CIFAR10, the learning rate for the gate variables are set to $0.001$ for $dir_2$ and $dir_3$, and set to $10^{-8}$ for $dir_1$. Here, a very small learning rate for $dir_1$ is used as the network has been pretrained and, thus, the gradients of the weights are expected to be small. Since $dir_1$ inverts the gradients of the weights in case the cost constraint is not satisfied, then the resulting direction will be large. Hence, a small learning rate is required to adjust the gate variables with reasonable steps. Similarly, the learning rate for $dir_2$ with layer gate variables, which is a single gate variable for the weights tensor of a layer, is set to $0.01$ as this yielded better results. Note that taking a very large learning rate for the gate variables results in a very fast reduction of the bit-widths, while a very small learning rate for the gate variables requires a very large number of epochs to obtain a sufficiently compressed model.

The output of the neural network is kept in floating point as was done by \cite{VanBaalen2020}. The input of the neural network is set to a fixed bit-width as this corresponds to the output of sensor data and is typically determined in advance and out of control of the network. In this work, this fixed bit-width of the input is set to 8-bit. Since the output is kept in floating point and the input is kept as a fixed bit-width, the activation of the output layer is not taken into account for the BOP count as they cannot be altered. As CGMQ can easily learn the different bit-widths each learnable parameter and activation will have a gate variable. This is different from BB where only a single bit-width is learned for each weight tensor or activation tensor. 

The gate variable initial value is set as $5.5$, which results in the model consisting of 32-bit weights at the start of training. No alternative initializations for the gate variables were performed in this work. In fact, all methods that learn bit-widths can use alternative initialization methods. Furthermore, no fine-tuning of the weights nor the quantization ranges are done after the learning cycle where the weights, the quantization ranges, and the bit-widths are learned together as this did not lead to better results.

Note that in the experimental section for each model a Relative Giga BOP (RGBOP) value is calculated by dividing the Giga BOP count of the quantized model by the giga BOP count of a model that uses for all weights and activations a 32-bit precision. Further remark that there exist a theoretical lower bound on the RGBOP which is achieved when all weights and activations are represented by only 2 bits. As indicated before pruning is considered outside of the scope of this work and is left as future work. In particular, the RGBOP for LeNet-5 is $0.392\%$ and for VGG-7 is $0.391\%$. On MNIST, only BB is considered as a baseline as it is the only work that provides BOP count and accuracy for the quantization of a LeNet-5 architecture with mixed bit-widths. It should be stressed that BB reported results for which pruning was active. On CIFAR10, both the BB and DQ are used as baselines for the same reasons. Also for this data set, pruning was active for the BB method. Next to a comparison with prior work, a set of experiments are done on the performance of CGMQ for different bounds on the cost.

\subsection{Results and discussion}

The results on MNIST with $B_{RGBOP}$, the upper bound on the relative Giga BOP count in $\%$, set to $0.4$ and different choices for $dir$ are shown in \cref{Tab:MNIST:Comparison}. First of all, all choices for the directions and the gate variables for CGMQ yield a model that satisfies the cost constraint. The best performing models of CGMQ are close to the performance of BB. Recall that BB does prune, which could explain a slightly higher performance as some variables can have larger bit-widths, because some others are pruned away while in CGMQ all variables have a bit-width of at least 2. It is remarkable that $dir_2$ is the worse performing choice for the direction, while $dir_1$ and $dir_3$ lead to models with a similar performance. Hence, we can conclude that on MNIST applying CGMQ results in models with a similar performance to BB with the advantage that CGMQ guarantees to find a model that satisfies the cost constraint.

\begin{table}[ht]
    \centering
    \begin{tabular}{l l c c c}
        \toprule
        Method & Hyperpar. & Acc & RGBOP & $B_{RGBOP}$ \\
        \midrule
        FP32 & $-$ & $99.36$ & $100$ & $100$ \\
        BB\tablefootnote{Results as reported in \citep{VanBaalen2020}.} & $\mu=0.01$ & $99.30 \pm 0.03$ & $0.36\pm0.01$ & - \\ \midrule
        FP32 & $-$ & $99.31$ & $100$ & $100$ \\
        CGMQ & $dir_1$, layer & $99.22$ & $0.39$ & $0.40$ \\
        CGMQ & $dir_2$, layer & $97.83$ & $0.39$ & $0.40$ \\
        CGMQ & $dir_3$, layer & $98.94$ & $0.40$ & $0.40$ \\
        CGMQ & $dir_1$, indiv. & $99.09$ & $0.39$ & $0.40$ \\
        CGMQ & $dir_2$, indiv. & $98.76$ & $0.40$ & $0.40$ \\
        CGMQ & $dir_3$, indiv. & $99.11$ & $0.40$ & $0.40$ \\
        \bottomrule 
    \end{tabular}
    \caption{Results on MNIST.}
    \label{Tab:MNIST:Comparison}
\end{table}

The results for the analogue experiment on CIFAR10 are shown in \cref{Tab:CIFAR10:Comparison}. Once again, CGMQ satisfies the cost constraint for all cases. However, $dir_1$ is the only choice for the direction that results in a well performing model compared to the baselines. The models with individual gate variables outperform their corresponding models with layer gate variables as is expected. The models obtained from BB are the best, however, these models prune some weights. This shows that extending CGMQ to allow to prune weights dynamically during training is important future work as it might hinder the current performance. Nevertheless, only a small difference in performance is observed. Hence, CGMQ achieves a slightly lower performance on CIFAR10 while a model that satisfies the cost constraint is guaranteed.

Next, the performance of CGMQ for different bounds in the cost constraint are considered. The results on MNIST for the layer gate variables and the individual gate variables are shown in \cref{Tab:MNIST:ParetoLayer} and \cref{Tab:MNIST:ParetoIndiv}, respectively. Once again, the obtained models satisfy the cost constraint. However, increasing the bound in the cost constraint does not always result in an improved performance. For $dir_1$, the increase in $B_{RGBOP}$ does not consistently result in a larger $RGBOP$ of the resulting network. This can be a consequence of the difference in obtained directions is too small to result in a significant difference. Hence, most bit-widths remain similar during training. Furthermore, $dir_2$ is less capable of finding more complex models when the bound in the cost constraint is increased. This illustrates that there is room for improvement on the choice of the directions as the performance should increase when more complex models are allowed.

\begin{table}[t]
    \centering
    \begin{tabular}{l l c c c}
        \toprule
        Method & Hyperpar. & Acc & RGBOP & $B_{RGBOP}$ \\
        \midrule
        FP32 & $-$ & $93.05$ & $100$ & $100$ \\
        DQ\tablefootnote{Results as reported in \citep{VanBaalen2020}.} & $-$ & $91.59$ & $0.48$ & - \\
        BB\footnote[2]{} & $\mu=0.01$ & $93.23 \pm 0.10$ & $0.51\pm 0.03$ & - \\
        BB\footnote[2]{} & $\mu=0.1$ & $91.96 \pm 0.04$ & $0.29\pm 0.00$ & - \\ \midrule
        FP32 & $-$ & $92.35$ & $100$ & $100$ \\
        CGMQ & $dir_1$, layer & $90.42$ & $0.54$ & $0.55$ \\
        CGMQ & $dir_2$, layer & $29.87$ & $0.52$ & $0.55$ \\
        CGMQ & $dir_3$, layer & $10.03$ & $0.45$ & $0.55$ \\
        CGMQ & $dir_1$, indiv. & $91.03$ & $0.54$ & $0.55$ \\
        CGMQ & $dir_2$, indiv. & $50.56$ & $0.52$ & $0.55$ \\
        CGMQ & $dir_3$, indiv. & $47.05$ & $0.53$ & $0.55$ \\
        \bottomrule
    \end{tabular}
    \caption{Results on CIFAR10.}
    \label{Tab:CIFAR10:Comparison}
\end{table}

\begin{table}[ht]
    \centering
    \begin{tabular}{l | c c c c c c c c}
        \toprule
        \multirow{2}{*}{$B_{RGBOP}$} & \multicolumn{2}{c}{$dir_1$ layer} & \multicolumn{2}{c}{$dir_2$ layer} & \multicolumn{2}{c}{$dir_3$ layer} \\
        & Acc & RGBOP & Acc & RGBOP & Acc & RGBOP \\
        \midrule
        $0.40$ & $99.22$ & $0.39$ & $97.83$ & $0.40$ & $98.94$ & $0.40$ \\
        $0.90$ & $99.31$ & $0.39$ & $97.35$ & $0.41$ & $99.14$ & $0.85$ \\
        $1.40$ & $99.21$ & $0.39$ & $97.35$ & $0.41$ & $99.04$ & $0.85$ \\
        $2.00$ & $99.12$ & $1.57$ & $97.35$ & $0.41$ & $99.21$ & $1.63$ \\
        $5.00$ & $99.30$ & $1.57$ & $97.35$ & $0.41$ & $99.27$ & $3.17$ \\
        \bottomrule 
    \end{tabular}
    \caption{Accuracy (Acc) in $\%$ and relative GBOPs (RGBOP) in $\%$ for different choices for the direction of the gradient variables and different bounds on the RGBOPs ($B_{RGBOP}$) in $\%$ of the neural network on MNIST for a single gate variable for the weights of a layer and a single gate variable for the activations of a layer.}
    \label{Tab:MNIST:ParetoLayer}
\end{table}

\begin{table}[ht]
    \centering
    \begin{tabular}{l | c c c c c c c c}
        \toprule
        \multirow{2}{*}{$B_{RGBOP}$} & \multicolumn{2}{c}{$dir_1$ indiv.} & \multicolumn{2}{c}{$dir_2$ indiv.} & \multicolumn{2}{c}{$dir_3$ indiv.} \\
        & Acc & RGBOP & Acc & RGBOP & Acc & RGBOP \\
        \midrule
        $0.40$ & $99.09$ & $0.39$ & $98.76$ & $0.40$ & $99.11$ & $0.40$ \\
        $0.90$ & $99.21$ & $0.39$ & $97.50$ & $0.56$ & $98.57$ & $0.90$ \\
        $1.40$ & $99.16$ & $0.39$ & $97.50$ & $0.56$ & $98.49$ & $1.30$ \\
        $2.00$ & $99.28$ & $1.56$ & $97.50$ & $0.56$ & $99.27$ & $2.00$ \\
        $5.00$ & $99.23$ & $1.56$ & $99.26$ & $5.00$ & $99.23$ & $5.00$ \\
        \bottomrule 
    \end{tabular}
    \caption{Accuracy (Acc) in $\%$ and relative GBOPs (RGBOP) in $\%$ for different choices for the direction of the gradient variables and different bounds on the RGBOPs ($B_{RGBOP}$) in $\%$ of the neural network on MNIST for a gate variable for each weight and activation individually.}
    \label{Tab:MNIST:ParetoIndiv}
\end{table}

The results for the analogue experiment on CIFAR10 for the layer gate variables and the individual gate variables are shown in \cref{Tab:CIFAR10:ParetoLayer} and \cref{Tab:CIFAR10:ParetoIndiv}, respectively. Similar to MNIST, when the model for the smallest $B_{RGBOP}$ achieves an accuracy of at least $90\%$, then an increase in complexity does not translate into an increase in performance. However, when the model for the smallest BGOP count has bad performance, then the performance increases with exception of $dir_2$ with layer gate variables where the performance remains constant. This is the expected trade-off between performance and complexity. A similar trade-off was observed on MNIST potentially due to some overfitting.

\begin{table}[ht]
    \centering
    \begin{tabular}{l | c c c c c c}
        \toprule
        \multirow{2}{*}{$B_{RGBOP}$} & \multicolumn{2}{c}{$dir_1$ layer} & \multicolumn{2}{c}{$dir_2$ layer} & \multicolumn{2}{c}{$dir_3$ layer} \\
        & Acc & RGBOP & Acc & RGBOP & Acc & RGBOP \\
        \midrule
        $0.55$ & $90.42$ & $0.54$ & $29.87$ & $0.52$ & $10.03$ & $0.45$ \\
        $1.40$ & $90.00$ & $1.09$ & $29.22$ & $1.07$ & $14.66$ & $0.91$ \\
        $5.00$ & $89.71$ & $3.33$ & $29.22$ & $1.07$ & $82.59$ & $3.45$ \\
        \bottomrule
    \end{tabular}
    \caption{Accuracy (Acc) in $\%$ and relative GBOPs (RGBOP) in $\%$ for different choices for the direction of the gradient variables and different bounds on the RGBOPs ($B_{RGBOP}$) in $\%$ of the neural network on CIFAR10 for a single gate variable for the weights of a layer and a single gate variable for the activations of a layer.}
    \label{Tab:CIFAR10:ParetoLayer}
\end{table}

\begin{table}[ht]
    \centering
    \begin{tabular}{l | c c c c c c}
        \toprule
        \multirow{2}{*}{$B_{RGBOP}$} & \multicolumn{2}{c}{$dir_1$ indiv.} & \multicolumn{2}{c}{$dir_2$ indiv.} & \multicolumn{2}{c}{$dir_3$ indiv.} \\
        & Acc & RGBOP & Acc & RGBOP & Acc & RGBOP \\
        \midrule
        $0.55$ & $91.03$ & $0.54$ & $50.56$ & $0.52$ & $47.05$ & $0.53$ \\
        $1.40$ & $91.59$ & $1.25$ & $84.36$ & $1.34$ & $83.11$ & $1.37$ \\
        $5.00$ & $91.54$ & $3.16$ & $90.78$ & $5.00$ & $90.56$ & $4.97$ \\
        \bottomrule
    \end{tabular}
    \caption{Accuracy (Acc) in $\%$ and relative GBOPs (RGBOP) in $\%$ for different choices for the direction of the gradient variables and different bounds on the RGBOPs ($B_{RGBOP}$) in $\%$ of the neural network on CIFAR10 for a gate variable for each weight and activation individually.}
    \label{Tab:CIFAR10:ParetoIndiv}
\end{table}

\section{Ablation study}\label{sec:AblationStudy}

An ablation study is performed on the choice for the bounds that convert the gate variable to a bit-width. In particular, we consider adjustment of the function $T$ as defined in Equation \ref{eq:DefT} as follows
    \begin{equation*}
        T^{a,b,c,d,e}: \mathbb{R}\to\mathbb{R}: g \mapsto \left\{ \begin{tabular}{l l}
            $0$, & if $g\leq a$, \\
            $2$, & if $g\in\left(a, b\right]$, \\
            $4$, & if $g\in\left(b, c\right]$, \\
            $8$, & if $g\in\left(c, d\right]$, \\
            $16$, & if $g\in\left(d, e\right]$, \\
            $32$, & if $g > e$.
        \end{tabular} \right.
    \end{equation*}
Observe that taking $a=0,b=1,c=2,d=3,e=4$ yields the original definition of $T$ as is shown in Equation \ref{eq:DefT}. In this work, the cases where $a=0,b=16,c=24,d=28,e=30$ and $a=0,b=2,c=4,d=4.5,e=5$ are considered. The reasoning behind the first choice is that decreasing a weight from 32-bit to 16-bit has a larger influence on the total BOP count of the model. Therefore, it might be more beneficial to allow weights to change quicker from 32-bit to 16-bit compared to switching from 8-bit to 4-bit. This is obtained by increasing the length of the intervals in the definition of $T^{a,b,c,d,e}$ accordingly. The reasoning behind the second choice is to investigate if small differences between the thresholds for the different bit-sizes lead to a bad performing model or not. The experiment on MNIST and CIFAR10 is repeated for one bound in the cost constraint. 

\subsection{Results and discussion}

The results for the ablation study on MNIST are shown in \cref{Tab:MNIST:Ablation:Ranges}. Here, no large difference can be observed between the different choices for the ranges as the largest difference is $0.2\%$. Therefore, we conclude that on MNIST the choice of ranges does not have a large influence on the resulting accuracy of the model on the test set.

\begin{table}[ht]
    \centering
    \begin{tabular}{l l l | r r r r r | c c c}
        \toprule
         & & & $a$ & $b$ & $c$ & $d$ & $e$ & Acc & RGBOP & $B_{RGBOP}$ \\
        \midrule
        \multirow{3}{*}{CGMQ} & \multirow{3}{*}{$dir_1$} & \multirow{3}{*}{indiv.} & $0$ & $1$ & $2$ & $3$ & $4$ & $99.17$ & $0.40$ & $1.50$ \\
        & & & $0$ & $16$ & $24$ & $28$ & $30$ & $99.07$ & $0.40$ & $1.50$ \\ 
        & & & $0$ & $2$ & $4$ & $4.5$ & $5$ & $99.27$ & $0.40$ & $1.50$ \\ 
        \midrule
        \multirow{3}{*}{CGMQ} & \multirow{3}{*}{$dir_1$} & \multirow{3}{*}{layer} & $0$ & $1$ & $2$ & $3$ & $4$ & $99.19$ & $0.40$ & $1.50$ \\
        & & & $0$ & $16$ & $24$ & $28$ & $30$ & $99.18$ & $0.40$ & $1.50$ \\ 
        & & & $0$ & $2$ & $4$ & $4.5$ & $5$ & $99.26$ & $0.40$ & $1.50$ \\ 
        \bottomrule
    \end{tabular}
    \caption{Accuracy (Acc) in $\%$ and relative GBOPs (RGBOP) in $\%$ for $dir_1$ and different ranges for the gate variables of the neural network on MNIST.}
    \label{Tab:MNIST:Ablation:Ranges}
\end{table}

The results of the ablation study on CIFAR10 are shown in \cref{Tab:CIFAR10:Ablation:Ranges}. Here, a large difference can be observed as the performance of $T^{0, 16,24,28,30}$ collapses. Hence, it appears that the ranges should be sufficiently close to each other and the relationship between the difference in two consecutive ranges with the improvement with respect to the total BOP count does not have a beneficial effect on the performance.

\begin{table}[ht]
    \centering
    \begin{tabular}{l l l | r r r r r | c c c}
        \toprule
         & & & $a$ & $b$ & $c$ & $d$ & $e$ & Acc & RGBOP & $B_{RGBOP}$ \\
         \midrule
        \multirow{3}{*}{CGMQ} & \multirow{3}{*}{$dir_1$} & \multirow{3}{*}{indiv.} & $0$ & $1$ & $2$ & $3$ & $4$ & $91.56$ & $1.49$ & $2.50$ \\
        & & & $0$ & $16$ & $24$ & $28$ & $30$ & $9.96$ & $0.39$ & $2.50$ \\ 
        & & & $0$ & $2$ & $4$ & $4.5$ & $5$ & $91.53$ & $1.41$ & $2.50$ \\ 
        \midrule
        \multirow{3}{*}{CGMQ} & \multirow{3}{*}{$dir_1$} & \multirow{3}{*}{layer} & $0$ & $1$ & $2$ & $3$ & $4$ & $89.39$ & $1.87$ & $2.50$ \\
        & & & $0$ & $16$ & $24$ & $28$ & $30$ & $9.96$ & $0.39$ & $2.50$ \\ 
        & & & $0$ & $2$ & $4$ & $4.5$ & $5$ & $89.74$ & $1.82$ & $2.50$ \\ 
        \bottomrule
    \end{tabular}
    \caption{Accuracy (Acc) in $\%$ and relative GBOPs (RGBOP) in $\%$ for $dir_1$ and different ranges for the gate variables of the neural network on CIFAR10.}
    \label{Tab:CIFAR10:Ablation:Ranges}
\end{table}

\section{Future work}\label{sec:FutureWork}
A valuable direction for further research is to investigate how other methods of quantization can be incorporated in the proposed method. For example, \cite{Uhlich2020} uses a different method to compute the quantization of a weight and activation which does yield a non-zero gradient for the bit-width. The current proposed method is build on the observation that the encoding of the quantization operation yields no gradient for the bit-width. Nevertheless, both could be combined by using the Constraint Guided Gradient Descent (CGGD) method \citep{VanBaelen2023}. Note that the gradients proposed in this work could be used as the direction of the constraints in CGGD. Due to the convergence properties of CGGD, the resulting method can yield similar convergence for the cost constraints. Similarly, using different estimators for the gradients in the backward pass than the STE estimator in this work might improve the resulting performance because it is known that the STE estimator has a large induced bias \citep{Wei2022,Shin2023}. However, for CGMQ as presented in this work, does not require the usage of STE for learning the gate variables $g$. The only requirement is that non-zero gradients can be obtained for the weights and the activations of the neural network.

The cost constraint which is formulated as an upper bound on the total BOP cost of the neural network served as a proof of concept. However, some hardware components are more efficient in handling certain bit-widths. Therefore, it would be valuable to consider a bit-width for group of parameters which are known to be combined during a forward pass and that determine the resulting speed of inference. By using this information, other constraints can be formulated that need to be satisfied such that the speed of inference of the resulting neural network is \textit{locally} optimal for a given hardware component. For example, considering the peak memory usage is important for the deployment on the edge. Hence, defining a constraint that defines an upper bound on the peak memory usage is crucial in future work.

Next, a detailed study on the properties of the directions that are used for learning the gate variables is crucial to find the best possible quantized network given some cost constraint. A possible new method for determining a direction could take into account the number of neurons in a layer as reducing the bit-width in a small layer might have a larger influence on the performance compared to the same action in a large layer. Furthermore, CGMQ should be generalized to support pruning such that even larger compression can be obtained, and, potentially, even better performance for the compression ratios considered in this work.

Since fine-tuning of weights and quantization ranges does result in an improvement in performance, it is crucial to investigate the influence of all steps in the quantization. In other words, CGMQ might not need to start from a pre-trained network. Note that initializing the gate variables with values that are large compared to the value $e$ of $T^{a,b,c,d,e}$ emulates quantization aware training for 32-bit parameters in the beginning of the training.

Finally, CGMQ should be tested on real-world applications. For example, it can be tested on larger networks and larger data sets such as the quantization of a ResNet18 \citep{He2016} on Imagenet. 

\section{Conclusion}\label{sec:Conclusion}

In this work, the Constraint Guided Model Quantization (CGMQ) algorithm is proposed such that a mixed precision neural network model can be learned that satisfied a predefined cost constraint. CGMQ combines the learning of the gate variables, which determine the bit-width of each weight, with the learning of the weights and the quantization ranges. Moreover, CGMQ does not require tuning of a hyperparameter that controls the trade-off between loss function and compression, which allows an easier usage of the method. In the experiments on MNIST and CIFAR10 it is indicated that CGMQ is competitive with respect to state-of-the-art methods. CGMQ supports different methods for learning the bit-widths as long as the sign is set correctly as a function of the satisfaction of the cost constraint. Therefore, future work should focus on investigating properties of these different methods and the study of desirable properties for large scale training sets and networks.

\section*{Acknowledgement}
This research received funding from the Flemish Government (AI ResearchProgram). This research has received support of Flanders Make. The authors declare that they have no known competing financial interests or personal relationships that could have appeared to influence the work reported in this paper.

\bibliographystyle{abbrvnat}
\bibliography{bibliographyarXiv}

\begin{thebibliography}{40}
\providecommand{\natexlab}[1]{#1}
\providecommand{\url}[1]{\texttt{#1}}
\expandafter\ifx\csname urlstyle\endcsname\relax
  \providecommand{\doi}[1]{doi: #1}\else
  \providecommand{\doi}{doi: \begingroup \urlstyle{rm}\Url}\fi

\bibitem[Baskin et~al.(2018)Baskin, Liss, Schwartz, Zheltonozhskii, Giryes,
  Bronstein, and Mendelson]{Baskin2018}
C.~Baskin, N.~Liss, E.~Schwartz, E.~Zheltonozhskii, R.~Giryes, A.~M. Bronstein,
  and A.~Mendelson.
\newblock Uniq: Uniform noise injection for non-uniform quantization of neural
  networks.
\newblock \emph{ACM Transactions on Computer Systems}, 37, 4 2018.
\newblock ISSN 15577333.
\newblock \doi{10.1145/3444943}.

\bibitem[Bengio et~al.(2013)Bengio, L{\'{e}}onard, and Courville]{Bengio2013}
Y.~Bengio, N.~L{\'{e}}onard, and A.~C. Courville.
\newblock Estimating or propagating gradients through stochastic neurons for
  conditional computation.
\newblock \emph{CoRR}, abs/1308.3432, 2013.

\bibitem[Bertsekas(2014)]{Bertsekas2014}
D.~Bertsekas.
\newblock \emph{Constrained Optimization and Lagrange Multiplier Methods}.
\newblock Computer science and applied mathematics. Elsevier Science, 2014.
\newblock ISBN 9781483260471.

\bibitem[Bulat and Tzimiropoulos(2021)]{Bulat2021}
A.~Bulat and G.~Tzimiropoulos.
\newblock Bit-mixer: Mixed-precision networks with runtime bit-width selection.
\newblock In \emph{Proceedings of the IEEE/CVF International Conference on
  Computer Vision}, pages 5188--5197, 2021.

\bibitem[Cai et~al.(2020)Cai, Yao, Dong, Gholami, Mahoney, and
  Keutzer]{Cai2020}
Y.~Cai, Z.~Yao, Z.~Dong, A.~Gholami, M.~W. Mahoney, and K.~Keutzer.
\newblock Zeroq: A novel zero shot quantization framework.
\newblock In \emph{Proceedings of the IEEE/CVF conference on computer vision
  and pattern recognition}, pages 13169--13178, 2020.

\bibitem[Chen et~al.(2021)Chen, Wang, and Cheng]{Chen2021}
W.~Chen, P.~Wang, and J.~Cheng.
\newblock Towards mixed-precision quantization of neural networks via
  constrained optimization.
\newblock In \emph{2021 IEEE/CVF International Conference on Computer Vision
  (ICCV)}, pages 5330--5339. IEEE, 10 2021.
\newblock ISBN 978-1-6654-2812-5.
\newblock \doi{10.1109/ICCV48922.2021.00530}.

\bibitem[Chu et~al.(2019)Chu, Luo, Yang, and Huang]{Chu2019}
T.~Chu, Q.~Luo, J.~Yang, and X.~Huang.
\newblock Mixed-precision quantized neural network with progressively
  decreasing bitwidth for image classification and object detection.
\newblock \emph{arXiv preprint arXiv:1912.12656}, 2019.

\bibitem[Eckstein and Bertsekas(1992)]{Eckstein1992}
J.~Eckstein and D.~P. Bertsekas.
\newblock On the douglas—rachford splitting method and the proximal point
  algorithm for maximal monotone operators.
\newblock \emph{Mathematical programming}, 55:\penalty0 293--318, 1992.

\bibitem[Elthakeb et~al.(2020)Elthakeb, Pilligundla, Mireshghallah,
  Yazdanbakhsh, and Esmaeilzadeh]{Elthakeb2020}
A.~T. Elthakeb, P.~Pilligundla, F.~Mireshghallah, A.~Yazdanbakhsh, and
  H.~Esmaeilzadeh.
\newblock Releq: A reinforcement learning approach for automatic deep
  quantization of neural networks.
\newblock \emph{IEEE micro}, 40\penalty0 (5):\penalty0 37--45, 2020.

\bibitem[Fish et~al.(2023)Fish, Michieli, and Ozay]{Fish2023}
E.~Fish, U.~Michieli, and M.~Ozay.
\newblock A model for every user and budget: Label-free and personalized
  mixed-precision quantization.
\newblock \emph{Proceedings of the Annual Conference of the International
  Speech Communication Association, INTERSPEECH}, 2023-August:\penalty0
  3232--3236, 7 2023.
\newblock ISSN 19909772.
\newblock \doi{10.21437/Interspeech.2023-61}.

\bibitem[Gabay and Mercier(1976)]{Gabay1976}
D.~Gabay and B.~Mercier.
\newblock A dual algorithm for the solution of nonlinear variational problems
  via finite element approximation.
\newblock \emph{Computers \& mathematics with applications}, 2\penalty0
  (1):\penalty0 17--40, 1976.

\bibitem[Gong et~al.(2019)Gong, Jiang, Wang, Lin, Liu, and Pan]{Gong2019}
C.~Gong, Z.~Jiang, D.~Wang, Y.~Lin, Q.~Liu, and D.~Z. Pan.
\newblock Mixed precision neural architecture search for energy efficient deep
  learning.
\newblock In \emph{2019 IEEE/ACM International Conference on Computer-Aided
  Design (ICCAD)}, pages 1--7. IEEE, 2019.

\bibitem[He et~al.(2016)He, Zhang, Ren, and Sun]{He2016}
K.~He, X.~Zhang, S.~Ren, and J.~Sun.
\newblock Deep residual learning for image recognition.
\newblock In \emph{Proceedings of the IEEE conference on computer vision and
  pattern recognition}, pages 770--778, 2016.

\bibitem[Hillar and Lim(2013)]{Hillar2013}
C.~J. Hillar and L.-H. Lim.
\newblock Most tensor problems are np-hard.
\newblock \emph{Journal of the ACM (JACM)}, 60\penalty0 (6):\penalty0 1--39,
  2013.

\bibitem[Hohman et~al.(2024)Hohman, Kery, Ren, and Moritz]{Hohman2024}
F.~Hohman, M.~B. Kery, D.~Ren, and D.~Moritz.
\newblock Model compression in practice: Lessons learned from practitioners
  creating on-device machine learning experiences.
\newblock In \emph{Proceedings of the CHI Conference on Human Factors in
  Computing Systems}, New York, NY, USA, 2024. Association for Computing
  Machinery.
\newblock ISBN 9798400703300.
\newblock \doi{10.1145/3613904.3642109}.

\bibitem[Jacob et~al.(2017)Jacob, Kligys, Chen, Zhu, Tang, Howard, Adam, and
  Kalenichenko]{Jacob2017}
B.~Jacob, S.~Kligys, B.~Chen, M.~Zhu, M.~Tang, A.~Howard, H.~Adam, and
  D.~Kalenichenko.
\newblock Quantization and training of neural networks for efficient
  integer-arithmetic-only inference.
\newblock \emph{Proceedings of the IEEE Computer Society Conference on Computer
  Vision and Pattern Recognition}, pages 2704--2713, 12 2017.
\newblock ISSN 10636919.
\newblock \doi{10.1109/CVPR.2018.00286}.

\bibitem[Kingma and Ba(2015)]{Kingma2015}
D.~P. Kingma and J.~L. Ba.
\newblock Adam: A method for stochastic optimization.
\newblock \emph{3rd International Conference on Learning Representations, ICLR
  2015 - Conference Track Proceedings}, pages 1--15, 12 2015.

\bibitem[Kingma and Welling(2013)]{Kingma2013}
D.~P. Kingma and M.~Welling.
\newblock Auto-encoding variational bayes.
\newblock \emph{2nd International Conference on Learning Representations, ICLR
  2014 - Conference Track Proceedings}, 12 2013.
\newblock \doi{10.61603/ceas.v2i1.33}.

\bibitem[Krishnamoorthi(2018)]{Krishnamoorthi2018}
R.~Krishnamoorthi.
\newblock Quantizing deep convolutional networks for efficient inference: A
  whitepaper.
\newblock \emph{arXiv preprint arXiv:1806.08342}, 2018.

\bibitem[Lacey et~al.(2018)Lacey, Taylor, and Areibi]{Lacey2018}
G.~Lacey, G.~W. Taylor, and S.~Areibi.
\newblock Stochastic layer-wise precision in deep neural networks.
\newblock \emph{34th Conference on Uncertainty in Artificial Intelligence 2018,
  UAI 2018}, 2:\penalty0 663--672, 7 2018.

\bibitem[Liu et~al.(2016)Liu, Li, Wang, Zhang, and Yan]{Liu2016}
B.~Liu, F.~Li, X.~Wang, B.~Zhang, and J.~Yan.
\newblock Ternary weight networks.
\newblock \emph{ICASSP, IEEE International Conference on Acoustics, Speech and
  Signal Processing - Proceedings}, 2023-June, 5 2016.
\newblock ISSN 15206149.
\newblock \doi{10.1109/ICASSP49357.2023.10094626}.

\bibitem[Lou et~al.(2019)Lou, Guo, Liu, Kim, and Jiang]{Lou2019}
Q.~Lou, F.~Guo, L.~Liu, M.~Kim, and L.~Jiang.
\newblock Autoq: Automated kernel-wise neural network quantization.
\newblock \emph{arXiv preprint arXiv:1902.05690}, 2019.

\bibitem[Menghani(2023)]{Menghani2023}
G.~Menghani.
\newblock Efficient deep learning: A survey on making deep learning models
  smaller, faster, and better.
\newblock \emph{ACM Computing Surveys}, 55:\penalty0 1--37, 12 2023.
\newblock ISSN 0360-0300.
\newblock \doi{10.1145/3578938}.

\bibitem[Nagel et~al.(2021)Nagel, Fournarakis, Amjad, Bondarenko, van Baalen,
  and Blankevoort]{Nagel2021}
M.~Nagel, M.~Fournarakis, R.~A. Amjad, Y.~Bondarenko, M.~van Baalen, and
  T.~Blankevoort.
\newblock A white paper on neural network quantization.
\newblock \emph{ArXiv}, abs/2106.08295, 2021.

\bibitem[Ning et~al.(2021)Ning, Chen, Zhang, and Shen]{Ning2021}
L.~Ning, G.~Chen, W.~Zhang, and X.~Shen.
\newblock Simple augmentation goes a long way: Adrl for dnn quantization.
\newblock In \emph{International Conference on Learning Representations}, 2021.

\bibitem[Pandey et~al.(2023)Pandey, Nagel, van Baalen, Huang, Patel, and
  Blankevoort]{Pandey2023}
N.~P. Pandey, M.~Nagel, M.~van Baalen, Y.~Huang, C.~Patel, and T.~Blankevoort.
\newblock A practical mixed precision algorithm for post-training quantization.
\newblock \emph{arXiv}, 2 2023.

\bibitem[Rakka et~al.(2024)Rakka, Fouda, Khargonekar, and Kurdahi]{Rakka2024}
M.~Rakka, M.~E. Fouda, P.~Khargonekar, and F.~Kurdahi.
\newblock A review of state-of-the-art mixed-precision neural network
  frameworks.
\newblock \emph{IEEE Transactions on Pattern Analysis and Machine
  Intelligence}, pages 1--20, 2024.
\newblock ISSN 0162-8828.
\newblock \doi{10.1109/TPAMI.2024.3394390}.

\bibitem[Shin et~al.(2023)Shin, So, Park, Kang, Yoo, and Park]{Shin2023}
J.~Shin, J.~So, S.~Park, S.~Kang, S.~Yoo, and E.~Park.
\newblock Nipq: Noise proxy-based integrated pseudo-quantization.
\newblock In \emph{Proceedings of the IEEE/CVF Conference on Computer Vision
  and Pattern Recognition (CVPR)}, pages 3852--3861, June 2023.

\bibitem[Singh and Gill(2023)]{Singh2023}
R.~Singh and S.~S. Gill.
\newblock Edge ai: A survey.
\newblock \emph{Internet of Things and Cyber-Physical Systems}, 3:\penalty0
  71--92, 1 2023.
\newblock ISSN 2667-3452.
\newblock \doi{10.1016/J.IOTCPS.2023.02.004}.

\bibitem[Uhlich et~al.(2020)Uhlich, Mauch, Cardinaux, Yoshiyama, Garcia,
  Tiedemann, Kemp, and Nakamura]{Uhlich2020}
S.~Uhlich, L.~Mauch, F.~Cardinaux, K.~Yoshiyama, J.~A. Garcia, S.~Tiedemann,
  T.~Kemp, and A.~Nakamura.
\newblock Mixed precision dnns: All you need is a good parametrization.
\newblock In \emph{International Conference on Learning Representations}, 2020.

\bibitem[van Baalen et~al.(2020)van Baalen, Louizos, Nagel, Amjad, Wang,
  Blankevoort, and Welling]{VanBaalen2020}
M.~van Baalen, C.~Louizos, M.~Nagel, R.~A. Amjad, Y.~Wang, T.~Blankevoort, and
  M.~Welling.
\newblock Bayesian bits: Unifying quantization and pruning.
\newblock In H.~Larochelle, M.~Ranzato, R.~Hadsell, M.~Balcan, and H.~Lin,
  editors, \emph{Advances in Neural Information Processing Systems}, volume~33,
  pages 5741--5752. Curran Associates, Inc., 2020.

\bibitem[{Van Baelen} and Karsmakers(2023)]{VanBaelen2023}
Q.~{Van Baelen} and P.~Karsmakers.
\newblock Constraint guided gradient descent: Training with inequality
  constraints with applications in regression and semantic segmentation.
\newblock \emph{Neurocomputing}, 556:\penalty0 126636, 2023.
\newblock ISSN 0925-2312.
\newblock \doi{https://doi.org/10.1016/j.neucom.2023.126636}.

\bibitem[Verhoef et~al.(2019)Verhoef, Laubeuf, Cosemans, Debacker, Papistas,
  Mallik, and Verkest]{Verhoef2019}
B.-E. Verhoef, N.~Laubeuf, S.~Cosemans, P.~Debacker, I.~Papistas, A.~Mallik,
  and D.~Verkest.
\newblock Fq-conv: Fully quantized convolution for efficient and accurate
  inference., 2019.

\bibitem[Wang et~al.(2019)Wang, Liu, Lin, Lin, and Han]{Wang2019}
K.~Wang, Z.~Liu, Y.~Lin, J.~Lin, and S.~Han.
\newblock Haq: Hardware-aware automated quantization with mixed precision.
\newblock In \emph{2019 IEEE/CVF Conference on Computer Vision and Pattern
  Recognition (CVPR)}, pages 8604--8612. IEEE, 6 2019.
\newblock ISBN 978-1-7281-3293-8.
\newblock \doi{10.1109/CVPR.2019.00881}.

\bibitem[Wei et~al.(2022)Wei, Gong, Li, Liu, and Yu]{Wei2022}
X.~Wei, R.~Gong, Y.~Li, X.~Liu, and F.~Yu.
\newblock Qdrop: Randomly dropping quantization for extremely low-bit
  post-training quantization.
\newblock \emph{ICLR 2022 - 10th International Conference on Learning
  Representations}, 3 2022.

\bibitem[Wu et~al.(2018)Wu, Wang, Zhang, Tian, Vajda, and Keutzer]{Wu2018}
B.~Wu, Y.~Wang, P.~Zhang, Y.~Tian, P.~Vajda, and K.~Keutzer.
\newblock Mixed precision quantization of convnets via differentiable neural
  architecture search.
\newblock \emph{arXiv preprint arXiv:1812.00090}, 2018.

\bibitem[Yao et~al.(2021)Yao, Dong, Zheng, Gholami, Yu, Tan, Wang, Huang, Wang,
  Mahoney, et~al.]{Yao2021}
Z.~Yao, Z.~Dong, Z.~Zheng, A.~Gholami, J.~Yu, E.~Tan, L.~Wang, Q.~Huang,
  Y.~Wang, M.~Mahoney, et~al.
\newblock Hawq-v3: Dyadic neural network quantization.
\newblock In \emph{International Conference on Machine Learning}, pages
  11875--11886. PMLR, 2021.

\bibitem[Yin et~al.(2022)Yin, Phan, Zang, Liao, and Yuan]{Yin2022}
M.~Yin, H.~Phan, X.~Zang, S.~Liao, and B.~Yuan.
\newblock Batude: Budget-aware neural network compression based on tucker
  decomposition.
\newblock \emph{Proceedings of the AAAI Conference on Artificial Intelligence},
  36:\penalty0 8874--8882, 6 2022.
\newblock ISSN 2374-3468.
\newblock \doi{10.1609/AAAI.V36I8.20869}.

\bibitem[Yuan et~al.(2020)Yuan, Chen, Hu, and Peng]{Yuan2020}
Y.~Yuan, C.~Chen, X.~Hu, and S.~Peng.
\newblock Evoq: Mixed precision quantization of dnns via sensitivity guided
  evolutionary search.
\newblock In \emph{2020 International Joint Conference on Neural Networks
  (IJCNN)}, pages 1--8. IEEE, 2020.

\bibitem[Zhe et~al.(2019)Zhe, Lin, Chandrasekhar, and Girod]{Zhe2019}
W.~Zhe, J.~Lin, V.~Chandrasekhar, and B.~Girod.
\newblock Optimizing the bit allocation for compression of weights and
  activations of deep neural networks.
\newblock In \emph{2019 IEEE International Conference on Image Processing
  (ICIP)}, pages 3826--3830. IEEE, 2019.

\end{thebibliography}

\end{document}